# A Multi-Biometrics for Twins Identification Based Speech and Ear

Cihan AKIN, Umit KACAR, Murvet KIRCI

*Department of Electrical and Electronic Engineer*
*Istanbul Technical University, Istanbul, Turkey*

**Abstract.** The development of technology biometrics becomes crucial more. To define human characteristic biometric systems are used but because of inability of traditional biometric systems to recognize twins, multimodal biometric systems are developed. In this study a multimodal biometric recognition system is proposed to recognize twins from each other and from the other people by using image and speech data. The speech or image data can be enough to recognize people from each other but twins cannot be distinguished with one of these data. Therefore a robust recognition system with the combine of speech and ear images is needed. As database, the photos and speech data of 39 twins are used. For speech recognition MFCC and DTW algorithms are used. Also, Gabor filter and DCVA algorithms are used for ear identification. Multi-biometrics success rate is increased by making matching score level fusion. Especially, rank-5 is reached 100%. We think that speech and ear can be complementary. Therefore, it is result that multi-biometrics based speech and ear is effective for human identifications.

## 1 Introduction

In 2009, 33.3 per 1000 births are twins in USA, 9-16 per 1000 births are twins in Europe and 18-30 per 1000 births are twins in Central Africa. With the increasing of twins rate in the world, new methods are needed to distinguish twins from each other. In the classical recognition systems, like face recognition, speech recognition, fingerprint recognition and ear recognition is not enough to distinguish twins from each other. Therefore, a multimodal recognition system is needed. In this study a multimodal recognition system for twins is developed based on ear recognition and speech recognition.

There are several database for multimodal recognition:

BANCA - is a European secure system project which is based on face and voice recognition in four European languages. In the database 208 people are used, 104 people are women and 104 people are men [1, 2].

MyIdea - is a multimodal biometric database designed to conduct research experiments in Identity Verification. It uses talking face, audio, fingerprints, signature, handwriting and hand geometry parameters. 104 people are used in database. French and English voice records are used [1, 3].

BioSec - is a multimodal working which has fingerprints images from three different sensors, frontal face images by webcam, iris images from an iris sensor and voice with a headset microphone and webcam microphone. It consist of 200 different subjects [1, 4].

In addition to these multimodal works there is also twins' multimodal biometric database.

Hamid Behravan and Karim Faez created a new database from twins' biometric traits. There are 6 biometric traits which are 2D face images, fingerprints, offline handwritten texts, videos of moving faces, spectral face images and thermal face images. Database consist of 104 subjects with 52 twins [1].

A face recognition based twins distinguishing project was worked in 2009-2010. The database consists of 150 pairs of twins, 300 subjects. 126 pairs' photos are taken in the same day and 24 pairs' photos taken one year apart [5].

CASIA database includes face, iris and fingerprint images 92 pairs of twins of 184 subject and 2 sets of triplets. Images are taken in same day and indoor. But not all subjects include in database. Database just includes subjects who have images face, 2 irises and 4 fingerprints. So there are 134 subjects, 64 pairs and 2 triplets [6, 7].

Audio-Visual Twins Database has 39 pairs of twins. Database contains face images, face videos and voice records of twins. Also twins have one right side photo and one left side photo which includes ear [6].

In this study Audio-Visual Twins Database is used. Voice records and side profile images are used. Because of lack of image left ear images are used as training set and right ear images are used as test set. Although database has 39 set of twins, one person has no right ear image, so 38 set of twins are used in workings. There are some problems in this working as written below:

- Twins
- Mirrored image
- One sample person
- Occlusion

- Rotation

## 2 A New Multi-biometrics

In the new modal voice records and ear images are used to distinguish twins from each other and from the other people. In the literature the combination of ear and voice is not much useful method.

Face images and audio records of Audio-Visual Twins Database are used but in this work multimodal ear images are needed. Therefore ear images are taken from side face images. In the database there are 39 sets of twins, 78 people. For speech recognition there are 3 different voice records for each person. Two of them are used as training and one of them is used test set. For ear recognition there is one left side face image and one right side face image. So two ear images are obtained from one of them left and one of them right side. One of them is used for training set and one of them is used for test set.

## 3 Speech Identification

Speech identification is one of the most useful method for biometric identify. Each person has a different speech characteristic but sometimes one person's speech characteristic is different in different times.

### 3.1. Feature Extracting

Speech recognition part of this system consists of two parts. First part is feature extraction and the second part is feature matching. For feature extraction MFCC (Mel Frequency Cepstral Coefficients) is used. MFCC is the mostly used algorithm to extract feature of speech signal. Because this algorithm models the human hearing perception so good. MFCC is basically shown below.

### 3.1.1 Framing-Windowing

Speech signal is more stable in the small size. So in signal processing systems, speech signal is divided into small size frames as 25-30ms. Signals are divided in to 30ms frames and the next frame begins after 10ms from the previous frame so there is an overlapping 25ms with two frames. Then, to destroy the quiet area of the speech signal, hamming window is applied to each frame.

$$framed\ signal : x(n)$$
$$hamming\ window : w(n) = 0.54 - 0.46 cos\left(\frac{2\pi n}{N-1}\right)$$
$$0 \leq n \leq N-1 \quad (1)$$
$$windowed\ signal : y(n) = w(n).x(n)$$

### 3.1.2 Fourier Transform

$$X(w) = \int_{-\infty}^{+\infty} x(t).e^{i2\pi ft} \quad (2)$$

By using Fast Fourier Transform, time domain coefficient is transformed to frequency domain coefficients.

### 3.1.3 Mel Frequency Filter Bank

Mel frequency is the best model of human's hearing. Mel scale is distributed linearly up to 1 kHz and after 1 kHz it changes logarithmic. The frequency domain coefficient of signal is passed from the mel scaled filter bank and taken the logarithm. Mel scaled filter bank has triangular filters to mel scale. The filter is chosen in the range of zero to half of the sampling frequency. 20 filter is settled in the range of 300Hz to 3700Hz.

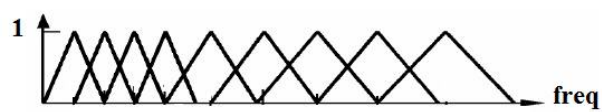

**Figure 1.** Mel Filter Bank.

### 3.1.4 Discrete Cosine Transform

Finally the data which logarithm is taken is transformed to time domain again by using DCT and Mel Frequency Cepstral Coefficients are obtained. 13 cepstral coefficients are obtained.

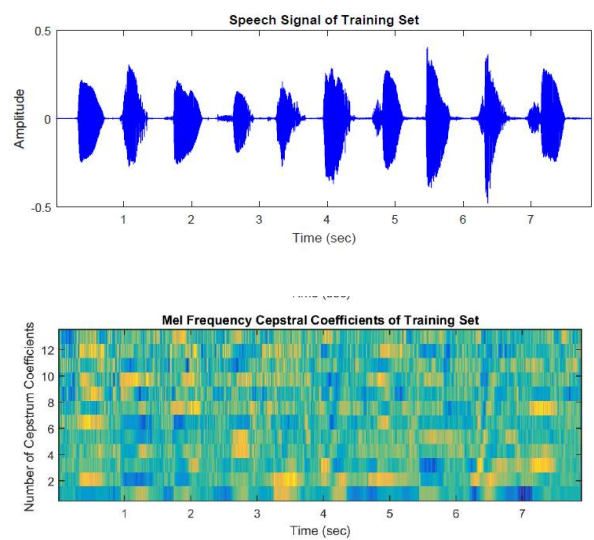

**Figure 2.** A speech signal and MFCC spectrum.

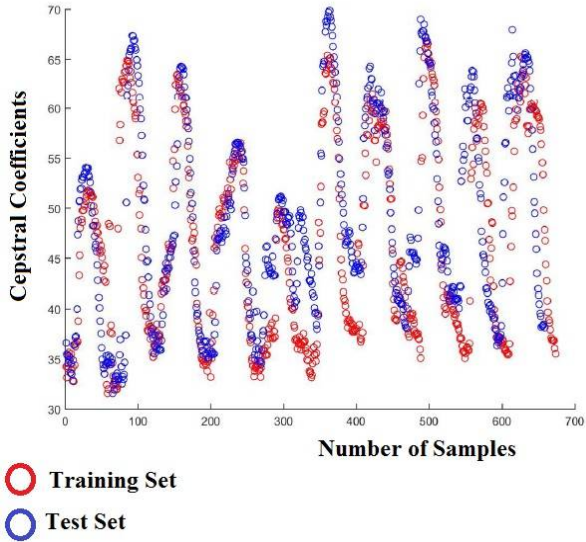

**Figure 3.** MFCC Scatter plot of the training and test set.

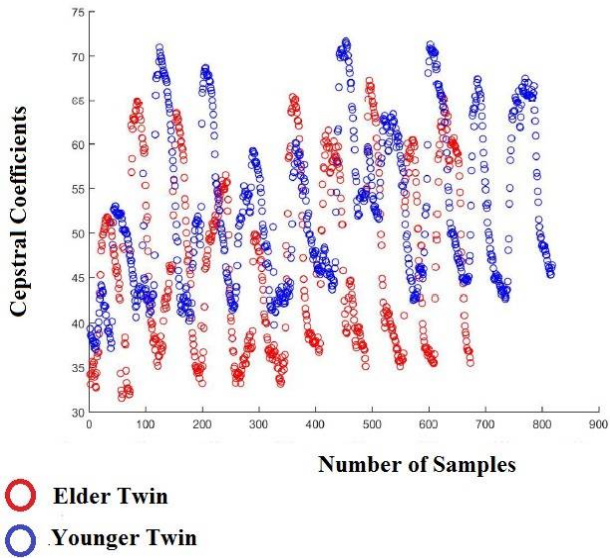

**Figure 4.** MFCC Scatter plot of the elder and younger twin.

### 3.2. Feature Comparision

Because of the people speak in different speed at the different times, although he/she says the same sentence, there will be time delay two of the sentences. To solve this problem Dynamic Time Warping (Doddington 1985) algorithm is used to find the similarities [9].

In DTW there are two vectors as training vector and test vector. These vectors' length does not need to be same length but in the working vectors are made equal.

DTW is the compares two vectors and find the minimum distance, as the result of the minimum distance the similarity of test vectors and training vectors is created. Training vector which is the best similar to training vector is chosen. DTW is formulated as below:

$$d(train_i, test_j) = \sqrt{(train_i - test_j)^2}$$
$$D(i,j) = \min[D(i-1,j-1), D(i-1,j), D(i,j-1)] + d(train_i, test_j) \quad (3)$$

## 4 Ear Identification

There is almost the study of twins' ear. The most known study among ear identification was made by Alfred Iannarelli at 1989, where he examined over 10.000 ears, he discovered that all ears were different even identical twins had similar but not identical ears [10]. There are many studies regarding ear recognition [11], even including embedded systems [12-13]. But, there is research gap in twins' ear. Our study includes both twins' ear (one-one) and all twins' ear (one all). However, Twins Ear Database have more problems: mirrored image, one sample person, occlusion and rotation etc.

### 4.1. Gabor Filter

Gabor filters were used to extract features from the cropped ear image in Twins Ear Database. Gabor filters is robust to rotation, translation, scale, illumination conditions, image noise etc. [14]. The Gabor filter-based features are separately extracted from the each R-G-B images.

2D Gabor filter is a Gaussian kernel function modulated by a complex sinusoidal plane wave, is shown below,

$$G(x,y) = \frac{f^2}{\pi\gamma\eta} \exp(-\frac{a^2 + (\gamma b)^2}{2a^2}) \exp(j2\pi fa + \psi)$$
$$a = x\cos\theta + y\sin\theta$$
$$b = -x\sin\theta + y\cos\theta \quad (4)$$

f is the frequency of the sinusoid, ψ is the phase offset, σ is the standard deviation of the Gaussian envelope, γ is the spatial aspect ratio which specifies the ellipticity of the support of the Gabor function and θ represents the orientation of the normal to the parallel stripes of a Gabor function.

320 Gabor filters in 32 orientations and 10 scales were used for feature extraction from Twins Ear Database. The size of the ear images is 360x280 pixels. Using 320 Gabor filters, the dimension of the feature vector is 360x280x320. It is reduced this data redundancy by downsampling the feature vectors resulting from Gabor filters owing to the adjacent pixels in an image are usually highly correlated. The feature vectors are downsampled by a factor of eight, so the dimension of feature vector is (360x280x320)/(8x8). Then, feature vectors which reduced dimension are normalized to zero mean and unit variance. Moreover, it is used Discriminative Common Vectors Approach (DCVA) to further reduce the size of the feature vectors.

### 4.2. Discriminative Common Vectors Approach

This method defines a common vector as discriminative to looking approaches in class. In addition to this these common vectors should create a projection which makes the best discriminative of classes. The features which are created by this projection are defined as the distinctive common vectors. As it is seen in equation (2), J is criteria function and ST is the in class covariance matrix of all classes [15, 16].

$$J_{DCVA}(W_{opt}) = \arg\max_{|W^T S_W W|=0} |W^T S_B W| = \arg\max_{|W^T S_B W|=0} |W^T S_T W| \quad (5)$$

A common matrix is created which defines every class. Therefore the usage of DCVA is more practical for reel time applications.

The dimension of feature vector using Gabor filter is (360x280x320)/(8x8). This feature vector is reduced to C-1 number of class by using DCVA. C is the number of class. Even if sample number in class increases, one feature vector is created which represents one class and identification comparison is realized with the feature vector.

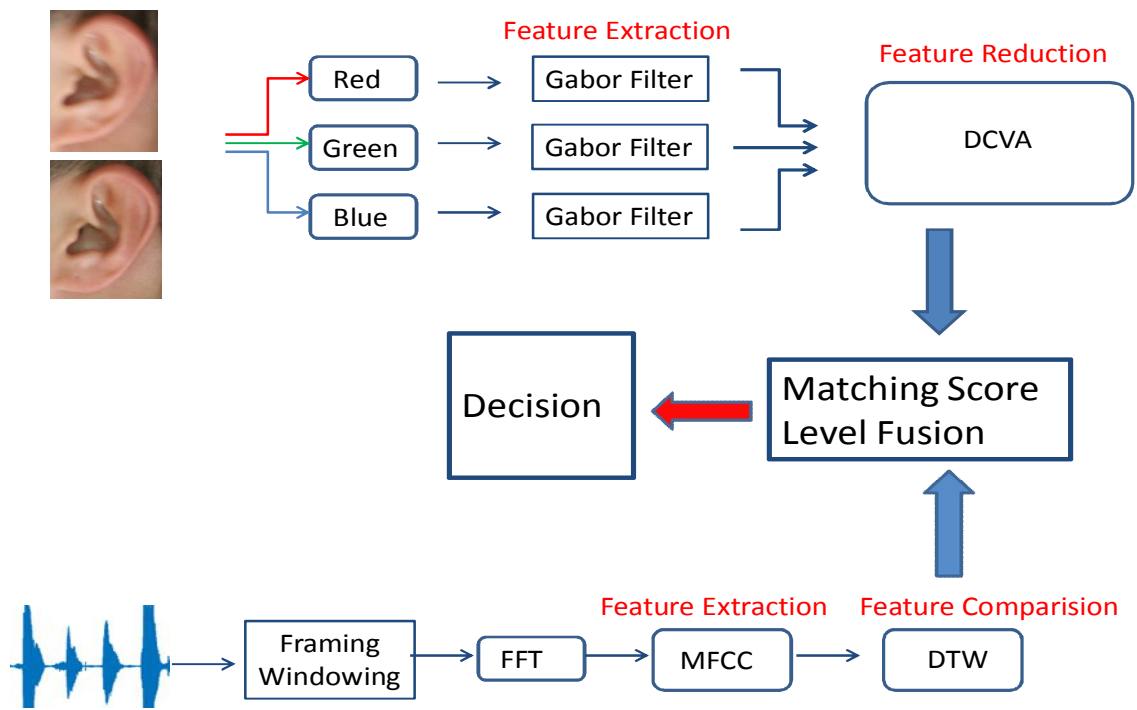

**Figure 5.** Block diagram of multi-biometrics system.

## 5. Experimental Results

MFCC result has 13 feature vector. By using 1 feature vector the speech identification success is 65% and then by using 4 feature vector of MFCC the success is increased to 80.3% as Cumulative Match Characteristic (CMC) curve is shown from the Fig.6. Moreover, only comparing twins' ear (one-one) success is 89.5%.

Gabor and DCVA finalized 75 feature vector which means C-1, for each one person. By using this feature vector the ear identification performance is 43.4%, as CMC curve is shown in Fig.7. Owing to below all problems, the performance is low. However, only comparing twins' ear (one-one) success is 72.4%.

- Twins
- Mirrored image
- One sample person
- Occlusion
- Rotation

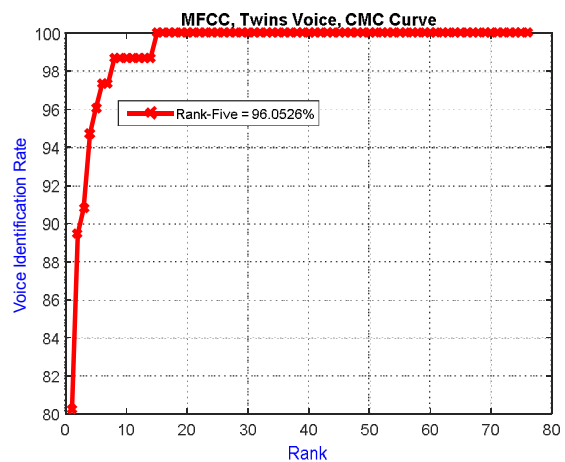

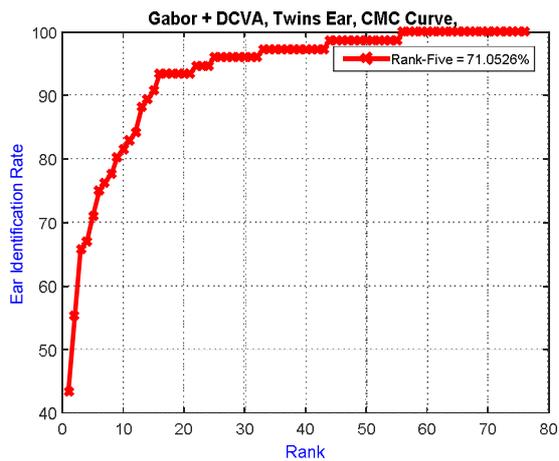

**Figure 6.** Speech identification.

**Figure 7.** Ear identification.

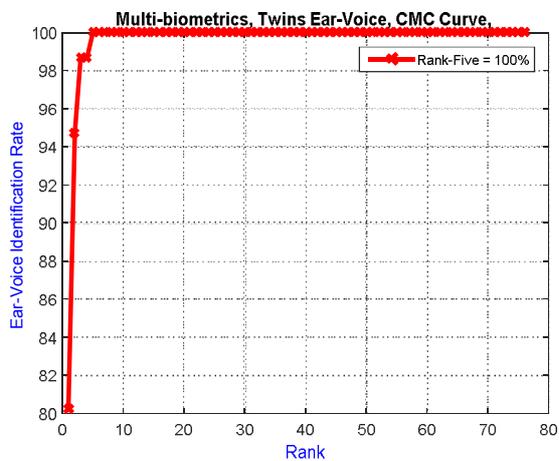

**Figure 8.** Multi-biometrics based speech-ear identification.

Speech and ear biometrics results are combined by matching score level fusion [17]. The matching scores cannot be used or combined directly; because these scores are based on different scaling techniques and from distinctive modalities. Score normalization is required, by changing over the scores into common similar scale or domain. Therefore, speech and ear matching scores was normalized between 0-1. Then, following weighted was applied. Due to the higher success rate of speech identification, speech was given more weighting.

Matching Score Fusion = 0.85*(speech) + 0.15*(ear)   (6)

As a result, total identification rate increases 81.6%, as CMC curve is shown in Fig.8 and only comparing twins' ear (one-one) success is 90.8% in rank-one. When other ranks are analyzed, especially rank-5 is reached 100%, as shown in Tab.1. Despite the low success of ear recognition, total success is reached 100% in rank-5. We think that speech and ear are mostly complementary. Therefore, it is result that multi-biometrics based speech and ear is effective for human identifications.

**Table 3.** The results of rank for multi-biometrics system.

| BIOMETRIC SYSTEM | Identification Rate % | | | | |
|---|---|---|---|---|---|
| | RANK-1 | RANK-2 | RANK-3 | RANK-4 | RANK-5 |
| Speech | 80.3 | 89.5 | 90.8 | 94.7 | 96.1 |
| Ear | 43.4 | 55.3 | 65.8 | 67.1 | 71.1 |
| Speech + Ear (matching score fusion) | 81.6 | 94.7 | 98.7 | 98.7 | **100** |
| Speech (only twins one –one) | 89.5 | | | | |
| Ear (only twins one –one) | 72.4 | | | | |
| Speech + Ear (only twins one –one) | **90.8** | | | | |

## 5. Conclusions

With the speech and ear data of twins by making matching score level fusion it was performed a multi-biometrics system. Twins Ear Database have more difficult problems than Twins Speech Database. Therefore, ear identification rate is lower than speech. However, identification rate is increased by making matching score level fusion. Especially, rank-5 is reached 100%. We think that speech and ear can be complementary. Therefore, it is result that multi-biometrics based speech and ear is effective for human identifications.

## References

1. Behravan, Hamid, and Karim Faez. "Introducing a new multimodal database from twins' biometric traits." *Electrical Engineering (ICEE), 2013 21st Iranian Conference on*. IEEE, 2013.
2. Bailly-Bailliére, E., Bengio, S., Bimbot, F., Hamouz, M., Kittler, J., Mariéthoz, J., ... & Ruiz, B. (2003, January). The BANCA database and evaluation protocol. In *Audio-and Video-Based Biometric Person Authentication* (pp. 625-638). Springer Berlin Heidelberg.
3. Fierrez, J., Ortega-Garcia, J., Toledano, D. T., & Gonzalez-Rodriguez, J. (2007). Biosec baseline corpus: A multimodal biometric database. *Pattern Recognition*, *40*(4), 1389-1392.
4. F. Alonso-Fernandez, J. Fierrez, D. Ramos and J. Ortega-Garcia, "BIOSECURE: A Multimodal Biometric Database", In Proceeding of SPIE, 2008.
5. Phillips, P. Jonathon, et al. "Distinguishing identical twins by face recognition."*Automatic Face & Gesture Recognition and Workshops (FG 2011), 2011 IEEE International Conference on*. IEEE, 2011.
6. Li, Jing, et al. "Audio-visual twins database." *Biometrics (ICB), 2015 International Conference on*. IEEE, 2015.
7. Sun, Z., Paulino, A. A., Feng, J., Chai, Z., Tan, T., & Jain, A. K. (2010, April). A study of multibiometric